\documentclass[10pt,twocolumn,letterpaper]{article}

\usepackage{btas}
\usepackage{times}
\usepackage{epsfig}
\usepackage{epstopdf}
\usepackage{graphicx}
\usepackage{amsmath}
\usepackage{amssymb}
\usepackage{enumerate}
\usepackage{subfigure}
\usepackage{tikz}
\usepackage{pgfplots}
\usepackage{overpic}
\newlength\figureheight 
\newlength\figurewidth 

\btasfinalcopy 

\ifbtasfinal\fi
\begin{document}

\title{ Generative Convolutional Networks for Latent Fingerprint Reconstruction}

\author{Jan Svoboda \textsuperscript{1}, Federico Monti \textsuperscript{1} and Michael M. Bronstein \textsuperscript{1, 2} \\
\textsuperscript{1}Institute of Computational Science, USI Lugano, Switzerland \\
\textsuperscript{2}Perceptual Computing, Intel, Israel \\
Email: {\tt\small \{jan.svoboda; federico.monti; michael.bronstein\}@usi.ch}
}

\maketitle
\thispagestyle{empty}

\begin{abstract}
Performance of fingerprint recognition depends heavily on the extraction of minutiae points. Enhancement of the fingerprint ridge pattern is thus an essential pre-processing step that noticeably reduces false positive and negative detection rates. A particularly challenging setting is when the fingerprint images are corrupted or partially missing. 
In this work, we apply generative convolutional networks to denoise visible minutiae and predict the missing parts of the ridge pattern. The proposed enhancement approach is tested as a pre-processing step in combination with several standard feature extraction methods such as MINDTCT, followed by biometric comparison using MCC and BOZORTH3. 
We evaluate our method on several publicly available latent fingerprint datasets captured using different sensors.

\end{abstract}

\section{Introduction}

Fingerprints have been in used as a means of person authentication since ancient time. The first use of fingerprints in a criminal investigation dates back to 1880's, when Francis Galton devised the first method for classifying fingerprints. In 1892, the first successful fingerprint-based identification helped convict a murderer. Since then, fingerprinting has underwent massive development and is nowadays an important part of crime scene investigation as well as modern biometric identification systems.

Based on their acquisition process, fingerprints can be categorized as inked, sensor-scan (e.g. optical, capacitive, etc.), or latent. 
The first two categories are typically further subdivided into rolled (nail-to-nail fingerprints), flat (single finger) or slap (four finger flat). Those have been heavily researched in the past yielding several state-of-the-art methods \cite{CLee01, Komarinsky04, Maltoni09}. 

\begin{figure}[!ht]
\centering
\subfigure[Synthetic Data]
{
\includegraphics[width=3.3in]{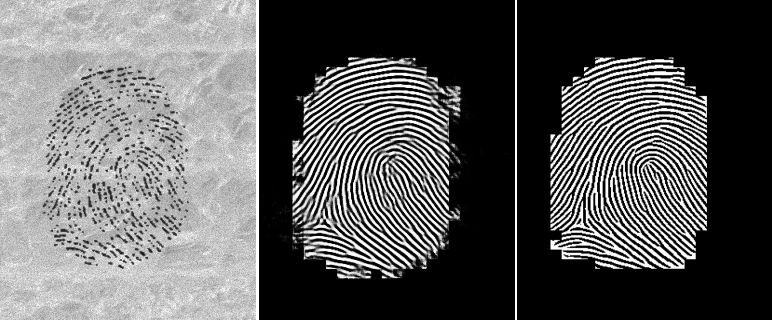}
}
\subfigure[Real Data]
{
\includegraphics[width=3.3in]{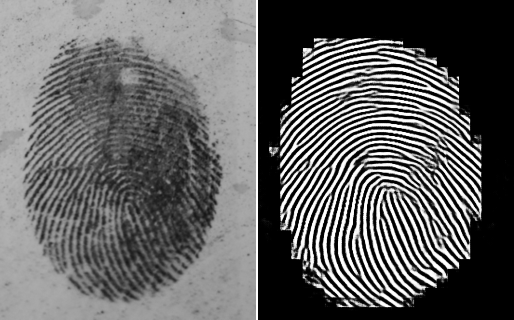}
}
\caption{Reconstruction of damaged or latent fingerprints using autoencoder networks. (a) Synthetic data example. From left to right: (input, reconstruction, target); (b) Real data - a pair (input, reconstruction). }
\label{fig:Example}
\end{figure}

On the other hand, latent fingerprints are impressions of the papillary lines that are unintentionally left by a subject at crime scenes. They are typically partial, blured, noisy, and exhibit poor ridge quality, as opposed to inked or sensor-scan fingerprints. Such fingerprints are usually lifted from the surface by means of special chemical procedures and photographed using high-resolution camera for further processing. 
Latent fingerprint matching is a challenging problem and state-of-the-art methods developed for inked and sensor-scan fingerprints do not work well (they typically fail to detect the minutiae or detect many false minutiae due to the imperfections mentioned above). 

In practice, latent fingerprints are analyzed with the help of forensic examiners who perform a manual latent fingerprint identification procedure called ACE-V (Analysis, Comparison, Evaluation, and Verification) \cite{Ashbaugh99}. This process is however very demanding and time consuming. In order to make identification efficient, forensic experts tend to restrict the population against which compare the latent fingerprints (focusing, for instance, only on suspects selected by witnesses or other evidence). This obviously reduces the likelihood of effectively identifying the culprit, thus making the overall process less reliable. 

Nowadays, several systems supporting the work of forensic examiners are available. The biggest one, AFIS (Automated Fingerprint Identification System), allows examiners to match latent fingerprints against large databases in a semi-automatic manner. The process usually consists of manually marking the minutiae points; launching the AFIS matcher; and visually verifying the top candidate fingerprints. 
Such a process requires considerable amount of tedious manual labour. 

In this paper, we improve latent fingerprint recognition by enhancing the input fingerprint image that allows using standard feature extraction methods more reliably. 
%
%
%
Inspired by the previous success of convolutional autoencoders (CAE) for image processing tasks \cite{Masci2011, ZhaoMGL15, MaoSY16a, NohHH15, HongNH15, DosovitskiySB14}, we design a convolutional autoencoder neural network capable at reconstructing high-quality fingerprint images from latent fingerprints. The neural network is trained on a synthetic dataset consisting of partial and blurry fingerprint impressions containing background noise; the output of the network is compared to the groundtruth fingerprint image. The training set was generated  using the open source implementation \cite{Ansari11} of  the SFinGE fingerprint generator \cite{Cappelli02}. 

We evaluate our method on two publicly available latent fingerprint datasets: IIIT-Delhi latent fingerprint \cite{Sankaran11} (for latent-to-latent fingerprints matching) and IIIT-Delhi MOLF  \cite{Sankaran15} (for latent-to-sensor-scan fingerprints matching). We show the broad applicability of our approach and discuss the possible future directions. All the evaluations were done using only open source software. We used MINDTCT \cite{NBIS} and an implementation of \cite{Abraham11} for features extraction, and BOZORTH3 \cite{NBIS} and MCC (Minutiae Cylinder Code) \cite{Cappelli10, Cappelli11, Ferrara12, Ferrara14} for fingerprints matching. We show that using our fingerprint enhancement method, the performance of non-commercial fingerprint matching algorithms can be improved and becomes comparable to some of the commercial ones.

The rest of the paper is organized as follows. In Section II, previous works are reviewed. Section III describes the objective function used for training the model. Section IV describes the convolutional autoencoder architecture. Section V presents the experimental results. Finally, Section VI concludes the paper and discusses possible future research directions.

\section{Related work}
Different solutions for improving latent fingerprint matching systems have been explored in the past. 
%
Many works have improved latent fingerprint matching by using extended features that require manual annotation \cite{JainFeng11, Yoon10}. However, performing the annotation in low-quality latent fingerprints may be very time-consuming and even infeasible in some settings. 
Other works have strived towards reducing the amount of manual input to only selection region of interest (ROI) and singular points \cite{Yoon10, Yoon11}. Approaches performing fusion of multiple matchers or multiple latent fingerprints were explored in \cite{Sankaran11}. 

A different direction for improving fingerprint matching concerns the enhancement of the acquired latent fingerprint images. Yoon \etal \cite{Yoon10} proposed a semi-automated method for enhancing the ridge information using the estimated orientation image. A more robust orientation field estimation technique based on Short-Time Fourier Transform and RANSAC has been proposed in \cite{Yoon11}. Feng \etal \cite{Feng12} proposed an approach capable of using prior knowledge on the fingerprint ridge structure employing a dictionary of reference orientated patches. Cao \etal \cite{Cao14} introduced a coarse-to-fine dictionary-based ridge enhancement technique.

Most recent work similar to the proposed approach is Schuch \etal \cite{Schuch16}, who used a convolutional autoencoder for inked and sensor-scan fingerprints ridge enhancement. Their network had a rather simple architecture and was not evaluated in challenging latent fingerprint recognition settings. 


\section{Gradient-based fingerprint similarity}
Since our method is based on Convolutional Neural Networks (CNN), the gradient analysis of the fingerprint image ridge pattern becomes a natural choice as the computation of the image gradient can be implemented using the convolutional operator.

\paragraph*{\bf Gradient filters. }

We compute the directional derivatives of the  fingerprint image $I$ by convolving it with the directional kernel $S_{\theta}$ 
\begin{equation}
G_{\theta}(I) = I \star S_{\theta},
\end{equation}
where $\theta$ is the direction in which we want to compute the gradient (we use 0, 45, 90, and 135 degrees).


As a criterion of similarity of two fingerprint images (target image $I_t$ and reconstructed image $I_r$), we used the 
average Mean Squared Error (MSE) 
on all the directions, 
\begin{equation}
E_{grad}(I_t, I_r) = \frac{\sum_{\theta \in \mathcal{T}} \frac{1}{n} \lVert (I_t - I_r)\star S_{\theta} \rVert_{2}^{2} }{\left\vert{\mathcal{T}}\right\vert},
\end{equation} 
where $\mathcal{T} = \{0, 45, 90, 135\}$ is the set of considered orientations and $n$ is the number of image pixels.
\paragraph*{\bf Ridge pattern orientation. }
Orientation of the ridge pattern can be defined through image moments using the already computed image gradients. Considering $G_x = G_{0}$ and $G_y = G_{90}$, we define the covariance matrix of the image using second order central moments ($\mu_{20}^{\prime}$, $\mu_{02}^{\prime}$, $\mu_{11}^{\prime}$) \cite{Hu62}:
\begin{equation}
\mathrm{Cov}(I(x, y)) = \begin{pmatrix}
\mu_{20}^{\prime} & \mu_{11}^{\prime} \\
\mu_{11}^{\prime} & \mu_{02}^{\prime} 
\end{pmatrix} = 
\begin{pmatrix}
G_{xx} & G_{xy} \\
G_{xy} & G_{yy}
\end{pmatrix},
\end{equation}
where $G_{xx} = g_{ \Sigma_s} \star (G_x \cdot G_x)$, $G_{yy} = g_{\Sigma_s} \star (G_y \cdot G_y)$ and $G_{xy} = g_{ \Sigma_s} \star (G_x \cdot G_y)$. $g_{\Sigma}$ represents a Gaussian smoothing kernel with covariance ${\Sigma}$ and $\cdot$ denotes the element-wise product.

Eigenvectors of the covariance matrix $\mathrm{Cov}(I(x, y))$ point in the directions of the major and minor intensity of image $I$. This information is enough to compute the orientation as the angle between the eigenvector corresponding to the largest eigenvalue and the x-axis is expressed by formula:
\begin{equation}
\Theta = \frac{1}{2} \tan^{-1} \left( \frac{2 G_{xy}}{G_{xx} - G_{yy}} \right).
\end{equation}

In order to strengthen the similarity between the reconstructed image and the associated ground-truth, we further calculate the {\em reliability orientation field} \cite{Khalil11}. The ridge orientation image is converted into a continuous vector field as 
\begin{equation}
(\Phi_x, \Phi_y) = (\cos(2 \Theta),  \sin(2 \Theta)),
\end{equation}
Subsequently, we apply Gaussian low-pass filter on the resulting vector field, 
\begin{equation}
(\Phi_x^{\prime}, \Phi_y^{\prime}) = (g_{\Sigma_o} \star \Phi_x,  g_{\Sigma_o} \star \Phi_y).
\end{equation}
The reliability measure $R$ is then defined by means of minimum inertia $I_{min}$ and maximum inertia $I_{max}$ as follows:
\begin{equation}
I_{min} = \frac{ (G_{yy} + G_{xx}) - (G_{xx} - G_{yy}) \Phi_x^{\prime} - G_{xy} \Phi_y^{\prime} }{2},
\end{equation}
\begin{equation}
I_{max} = G_{yy} + G_{xx } - I_{min},
\end{equation}
\begin{equation}
R = 1 - \frac{I_{min}}{I_{max}}.
\end{equation}
The intuition behind is that when the ratio $I_{min}/I_{max}$ close to 1, there is very little orientation information at that point.

Finally, we define the orientation energy $E_{ori}$ and reliability energy $E_{rel}$ as the MSE of the orientation and reliability measures computed on the target image $I_t$ and the reconstructed image $I_r$,
\begin{equation}
E_{ori} = \frac{1}{n} \lVert \Theta(I_t) - \Theta(I_r) \rVert_{2}^{2},
\end{equation}
\begin{equation}
E_{rel} = \frac{1}{n} \lVert R(I_t) - R(I_r) \rVert_{2}^{2}.
\end{equation}

\section{Fingerprint reconstruction model}

\begin{figure*}
 \label{fig:arch}
  \includegraphics[width=\textwidth]{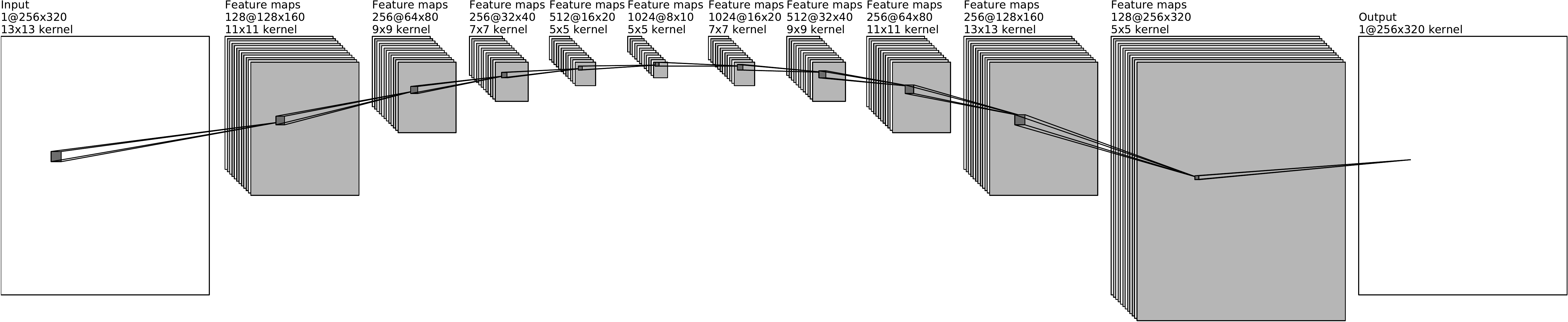}
  \caption{Our fully convolutional autoencoder.}
\end{figure*}

Following the principles described in the previous sections and the guidelines presented in \cite{RadfordMC15}, we approached the fingerprint reconstruction problem using a fully convolutional autencoder network.


\paragraph*{\bf Architecture. }

The encoding part corresponds to a fully convolutional neural network comprising five convolutional layers. It receives as input tensors of dimensions 1@260x320 and produces as output tensors of dimensions 1024@8x10 (F@WxH is a compact notation for tensors with width W, height H and F different feature channels). The convolutional layers have stride equal to 2 and are equipped with REctified Linear Unit (RELU) \cite{icml2010_NairH10}. Batch normalization is applied to each layer for faster convergence \cite{IoffeS15}. 

The output of the encoder is directly fed as an input to the decoding part. The decoder copies the architecture of the encoder performing the following changes. Each convolutional layer is replaced with de-convolutional layer (\ie a fractionally-strided convolutional layer with stride equal to 0.5 \cite{RadfordMC15}), and each RELU with a Leaky-RELU \cite{maas2013rectifier}. In order to reproduce the original grey scale image, a further convolutional layer equipped with a sigmoid activation function is applied at the end of the decoder. Figure 2 depicts our architecture.

\paragraph*{\bf Training objective. }
Based on the analysis described in Section 3, we define an objective function that allows to efficiently train our convolutional autoencoder to reconstruct corrupted parts of the fingerprint images. The objective function corresponds to a weighted average of three different losses:
\begin{equation}
E = E_{grad} + \lambda ( E_{ori} + E_{rel}),
\label{eq:Objective}
\end{equation}
where $E_{grad}$, $E_{ori}$ and $E_{rel}$ are as defined in the previous section, and $\lambda$ is a parameter weighting the contribution of the orientation and reliability regularizers. 

\paragraph*{\bf Training. }
For the purpose of training our model, we generated a dataset of synthetic fingerprints. From these fingerprint images, we simulated latent fingerprint images by applying rotation, translation,  directional blur and morphological dilation, and blend the results with several different backgrounds. 
%
This way, we aim to simulate the latent fingerprint formation as reliably as possible. We binarize the groudtruth (target) images using approach based on \cite{Bartunek06}. 
The network is applied to the synthetic latent fingerprint images and tries to reconstruct the underlying groundtruth image by minimizing the above objective function between the output and the groundtruth image. 
Since the groundtruth images are binarized, our model learns this way not only to reconstruct the fingerprints, but also to directly produce binary images.

Overall, we used 15000 synthetic fingerprint images for training. Data augmentation was performed by adding random Gaussian i.i.d. noise with $\mu = 0$ and $\sigma = 3.5 \cdot 10^{-3}$. 
%
The network was trained for 400 epochs (each epoch consisting of 64 iterations). In each iteration, we fed the network with a batch of 12 latent/groundtruth image pairs. 
We employed Adam \cite{KingmaB14} updates with $\beta_1 = 0.5$ and weight decay using $L_2$ regularization with $\mu = 10^{-4}$. The learning rate is set to $2 \cdot 10^{-4}$. The weighting parameter $\lambda$ was set to $0.1$.

\section{Experiments}\label{sec:experiments}
We demonstrate the proposed method on several settings of latent fingerprint recognition.
%
The experiments were carried out using publicly available datasets. Latent fingerprint enhancement was evaluated using the IIIT-Delhi Latent fingerprint \cite{Sankaran11} and IIIT-Delhi MOLF \cite{Sankaran15} datasets. 

For the first evaluation, we applies standard fingerprint recognition algorithms on original fingerprint images and those enhanced by our method. We employed two different feature extraction methods: ABR11 proposed by Abraham \etal \cite{Abraham11} and MINDTCT from NBIS \cite{NBIS}. Extracted features were subsequently compared using two different methods, BOZORTH3 \cite{NBIS} (abbreviated to BOZ in the following) and Minutia Cylinder Code (MCC) \cite{Cappelli10, Cappelli11, Ferrara12, Ferrara14}.
For the other evaluations, we used the combination ABR11 + MCC as the best performing one for latent fingerprint matching.
Results of latent fingerprint enhancement were presented using TopX measure and Cumulative Match Characteristic (CMC) curves. 

\paragraph*{\bf Latent-to-Latent matching. }
We evaluated latent-tolatent fingerprint matching using the protocol described in \cite{Sankaran11}. The whole IIIT-Delhi latent fingerprint dataset contained 1046 samples of all the ten fingerprints collected from 15 different subjects. We followed the dataset split strategy proposed in \cite{Sankaran11}, randomly choosing 395 images as gallery and 520 as probes, making sure that each class contained at least one gallery sample. 131 images were left out since Sankaran \etal used them for training. We performed 10-fold cross-validation in order to ensure the random splitting does not influence the reported performance. 

Table \ref{tab:LatentToLatent}, shows Rank-1 and Rank-10 accuracy for both recognition methods with and without our enhancement. Our approach significantly improves the matching accuracy. Comparing to \cite{Sankaran11}, we outperform all the fingerprint matching methods they have evaluated. It is worth emphasizing that differently from Sankaran \etal we do not need to train on the subset of the data as they do. 

\begin{figure}[!t]
\centering
\setlength\figureheight{6cm} 
\setlength\figurewidth{\linewidth}
%
%
%
\begin{tikzpicture}

\pgfplotsset{compat=newest} 

\definecolor{color0}{rgb}{1, 1, 1}

\tikzstyle{every node}=[font=\footnotesize]

\begin{axis}[
xlabel={Rank-N},
ylabel={Accuracy (\%)},
xmin=1, xmax=25,
ymin=20, ymax=100,
width=\figurewidth,
height=\figureheight,
at={(0\figurewidth,0\figureheight)},
xmajorgrids,
x grid style={lightgray},
ymajorgrids,
y grid style={lightgray},
axis line style={black},
axis background/.style={fill=color0},
legend style={at={(0.97,0.03)}, anchor=south east},
legend cell align={left},
legend entries={{MINDTCT+BOZ},{ABR11+MCC},{OUR+MINDTCT+BOZ},{OUR+ABR11+MCC}}
]
\addplot [dashed, line width=0.75pt, green]
table {%
    1.0000   57.6923
    2.0000   61.3462
    3.0000   63.4615
    4.0000   65.5769
    5.0000   66.9231
    6.0000   68.6538
    7.0000   70.1923
    8.0000   70.3846
    9.0000   71.3462
   10.0000   71.9231
   11.0000   72.8846
   12.0000   73.6538
   13.0000   74.0385
   14.0000   74.2308
   15.0000   74.8077
   16.0000   75.5769
   17.0000   76.5385
   18.0000   77.1154
   19.0000   77.1154
   20.0000   77.6923
   21.0000   77.8846
   22.0000   78.2692
   23.0000   78.2692
   24.0000   78.6538
   25.0000   78.6538
};
\addplot [dashed, line width=0.75pt, red]
table {%
    1.0000   35.5769
    2.0000   42.1154
    3.0000   45.5769
    4.0000   48.4615
    5.0000   51.1538
    6.0000   51.9231
    7.0000   53.2692
    8.0000   55.0000
    9.0000   56.3462
   10.0000   58.2692
   11.0000   59.0385
   12.0000   60.3846
   13.0000   61.1538
   14.0000   61.7308
   15.0000   62.5000
   16.0000   63.0769
   17.0000   64.8077
   18.0000   65.1923
   19.0000   65.7692
   20.0000   66.7308
   21.0000   66.9231
   22.0000   67.3077
   23.0000   67.6923
   24.0000   67.8846
   25.0000   68.4615
};
\addplot [line width=1.25, green!50.0!black]
table {%
    1.0000   62.6923
    2.0000   67.6923
    3.0000   70.3846
    4.0000   72.5000
    5.0000   74.0385
    6.0000   74.8077
    7.0000   75.9615
    8.0000   77.6923
    9.0000   78.4615
   10.0000   78.8462
   11.0000   79.4231
   12.0000   80.1923
   13.0000   80.3846
   14.0000   81.1538
   15.0000   82.1154
   16.0000   82.5000
   17.0000   82.8846
   18.0000   82.8846
   19.0000   83.6538
   20.0000   84.0385
   21.0000   84.8077
   22.0000   85.3846
   23.0000   85.9615
   24.0000   86.1538
   25.0000   87.1154
};
\addplot [line width=1.25, red]
table {%
    1.0000   57.1154
    2.0000   63.0769
    3.0000   67.5000
    4.0000   70.1923
    5.0000   72.3077
    6.0000   73.8462
    7.0000   75.9615
    8.0000   76.5385
    9.0000   77.8846
   10.0000   79.0385
   11.0000   80.1923
   12.0000   80.9615
   13.0000   81.5385
   14.0000   81.7308
   15.0000   82.1154
   16.0000   83.2692
   17.0000   84.0385
   18.0000   84.2308
   19.0000   84.4231
   20.0000   85.1923
   21.0000   85.5769
   22.0000   85.9615
   23.0000   86.1538
   24.0000   86.5385
   25.0000   86.5385
};
\end{axis}

\end{tikzpicture}
\vspace{-5mm}
\caption{ Latent-to-Latent matching on IIIT-Delhi latent database. CMC curve showing the performance improvement of our enhancement approach (solid lines) against matching raw latent fingerprints (dashed lines). }
\label{fig:cmcLatentToLatent}
\end{figure}
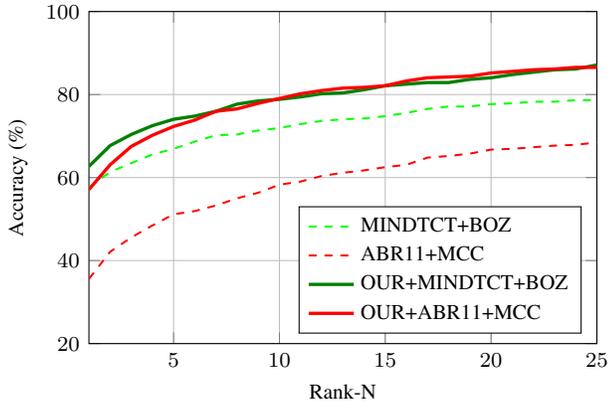

The CMC curves for this experiment are shown in Figure \ref{fig:cmcLatentToLatent}. Our model boosts the performance of ABR11 + MCC much more than that of NBIS. We attribute this to the fact that the energy we minimize while training our model performs very similar operations as the ridge binarization part of the ABR11 feature extraction algorithm. 


\paragraph*{\bf Latent-to-Sensor matching. }
Our second set of experiments tackled an even more challenging task. We used the MOLF dataset containing all ten  fingerprints of 100 different subjects. The samples in this dataset were of very different quality, including some very poor samples where no ridge structure was visible at all. Fingerprints of each participant were captured with several commercial fingerprint scanners (Lumidigm, Secugen and Crossmatch). In addition, each participant provided a set of latent fingerprints. It is therefore possible to match latent fingerprints to those acquired by a sensor. Following the testing protocol by Sankaran \etal \cite{Sankaran15}, we considered the first and second instance fingerprints for each user from a sensor scanned database as the gallery. The whole latent fingerprint database consisting of 4400 samples was considered as probe set. 

\begin{figure}[!t]
\centering
\setlength\figureheight{6cm} 
\setlength\figurewidth{\linewidth}
%
%
%
\begin{tikzpicture}

\pgfplotsset{compat=newest} 

\definecolor{color0}{rgb}{1, 1, 1}

\tikzstyle{every node}=[font=\footnotesize]

\begin{axis}[
xlabel={Rank-N},
ylabel={Accuracy (\%)},
xmin=1, xmax=50,
ymin=0, ymax=30,
width=\figurewidth,
height=\figureheight,
at={(0\figurewidth,0\figureheight)},
xmajorgrids,
x grid style={lightgray},
ymajorgrids,
y grid style={lightgray},
axis line style={black},
axis background/.style={fill=color0},
legend style={at={(0.03,0.97)}, anchor=north west},
legend cell align={left},
legend entries={{MINDTCT+MCC},{ABR11+MCC},{OUR+MINDTCT+MCC},{OUR+ABR11+MCC}}
]
\addplot [dashed, line width=0.75pt, blue]
table {%
    1.0000    1.2727
    2.0000    2.0455
    3.0000    2.6136
    4.0000    3.1591
    5.0000    3.3636
    6.0000    3.7273
    7.0000    4.0000
    8.0000    4.2273
    9.0000    4.4545
   10.0000    4.7273
   11.0000    4.9318
   12.0000    5.2273
   13.0000    5.3636
   14.0000    5.6136
   15.0000    6.0227
   16.0000    6.2045
   17.0000    6.3864
   18.0000    6.5909
   19.0000    6.8409
   20.0000    7.1591
   21.0000    7.2955
   22.0000    7.4545
   23.0000    7.7045
   24.0000    7.8636
   25.0000    8.0227
   26.0000    8.1364
   27.0000    8.3409
   28.0000    8.5682
   29.0000    8.6364
   30.0000    8.8409
   31.0000    9.0455
   32.0000    9.1818
   33.0000    9.4091
   34.0000    9.5227
   35.0000    9.7045
   36.0000    9.8409
   37.0000   10.0909
   38.0000   10.2500
   39.0000   10.4318
   40.0000   10.5455
   41.0000   10.7955
   42.0000   10.9545
   43.0000   11.1136
   44.0000   11.2955
   45.0000   11.5455
   46.0000   11.7500
   47.0000   11.8864
   48.0000   12.0909
   49.0000   12.3409
   50.0000   12.5909
};
\addplot [dashed, line width=0.75pt, red]
table {%
    1.0000    0.1818
    2.0000    0.3864
    3.0000    0.6136
    4.0000    0.6818
    5.0000    0.7727
    6.0000    0.8864
    7.0000    0.9545
    8.0000    1.0455
    9.0000    1.1136
   10.0000    1.1818
   11.0000    1.3182
   12.0000    1.5455
   13.0000    1.5909
   14.0000    1.7727
   15.0000    1.9318
   16.0000    2.0227
   17.0000    2.1591
   18.0000    2.2955
   19.0000    2.3864
   20.0000    2.4545
   21.0000    2.5909
   22.0000    2.7500
   23.0000    2.8409
   24.0000    3.0227
   25.0000    3.0682
   26.0000    3.1364
   27.0000    3.1818
   28.0000    3.3409
   29.0000    3.3636
   30.0000    3.5000
   31.0000    3.6136
   32.0000    3.7273
   33.0000    3.8636
   34.0000    3.9091
   35.0000    4.0909
   36.0000    4.1818
   37.0000    4.2727
   38.0000    4.4091
   39.0000    4.6136
   40.0000    4.6818
   41.0000    4.7955
   42.0000    4.8864
   43.0000    5.0000
   44.0000    5.0455
   45.0000    5.1591
   46.0000    5.1591
   47.0000    5.1818
   48.0000    5.2500
   49.0000    5.3182
   50.0000    5.4545
};
\addplot [line width=1.25, blue]
table {%
    1.0000    2.1136
    2.0000    3.0455
    3.0000    3.8409
    4.0000    4.7500
    5.0000    5.2955
    6.0000    5.8182
    7.0000    6.2500
    8.0000    6.7955
    9.0000    7.1364
   10.0000    7.6591
   11.0000    8.0000
   12.0000    8.3182
   13.0000    8.7500
   14.0000    9.1364
   15.0000    9.3636
   16.0000    9.6364
   17.0000   10.0000
   18.0000   10.2500
   19.0000   10.5682
   20.0000   10.8864
   21.0000   11.2045
   22.0000   11.5455
   23.0000   11.7955
   24.0000   12.0909
   25.0000   12.5455
   26.0000   12.7727
   27.0000   13.1591
   28.0000   13.3409
   29.0000   13.7500
   30.0000   13.9773
   31.0000   14.2045
   32.0000   14.4773
   33.0000   14.5682
   34.0000   14.8182
   35.0000   14.9773
   36.0000   15.3636
   37.0000   15.5909
   38.0000   15.7500
   39.0000   15.9318
   40.0000   16.1591
   41.0000   16.3864
   42.0000   16.6136
   43.0000   16.8864
   44.0000   17.1136
   45.0000   17.2955
   46.0000   17.5682
   47.0000   17.7727
   48.0000   18.0682
   49.0000   18.2727
   50.0000   18.3636
};
\addplot [line width=1.25, red]
table {%
    1.0000    3.4773
    2.0000    5.0000
    3.0000    6.1818
    4.0000    7.0000
    5.0000    7.7500
    6.0000    8.3409
    7.0000    8.8636
    8.0000    9.4091
    9.0000   10.1364
   10.0000   10.4091
   11.0000   10.8182
   12.0000   11.2273
   13.0000   11.5909
   14.0000   11.9773
   15.0000   12.4318
   16.0000   12.7273
   17.0000   13.0682
   18.0000   13.7273
   19.0000   14.1591
   20.0000   14.6818
   21.0000   15.1136
   22.0000   15.3864
   23.0000   15.8182
   24.0000   15.9773
   25.0000   16.1364
   26.0000   16.4545
   27.0000   16.7727
   28.0000   17.0682
   29.0000   17.4091
   30.0000   17.7273
   31.0000   18.0682
   32.0000   18.2500
   33.0000   18.6818
   34.0000   18.9545
   35.0000   19.2273
   36.0000   19.4091
   37.0000   19.7045
   38.0000   19.8409
   39.0000   20.0682
   40.0000   20.3636
   41.0000   20.5227
   42.0000   20.7500
   43.0000   20.8864
   44.0000   21.0227
   45.0000   21.2727
   46.0000   21.4773
   47.0000   21.7500
   48.0000   21.9318
   49.0000   22.1364
   50.0000   22.3636
};
\end{axis}

\end{tikzpicture}
\caption{Latent-to-Sensor matching on IIIT-Delhi MOLF database. CMC curves showing comparison of our enhancement approach (solid lines) against matching raw latent fingerprints (dashed lines) to the Lumidigm database. }
\label{fig:cmcLatentToSensor}
\end{figure}
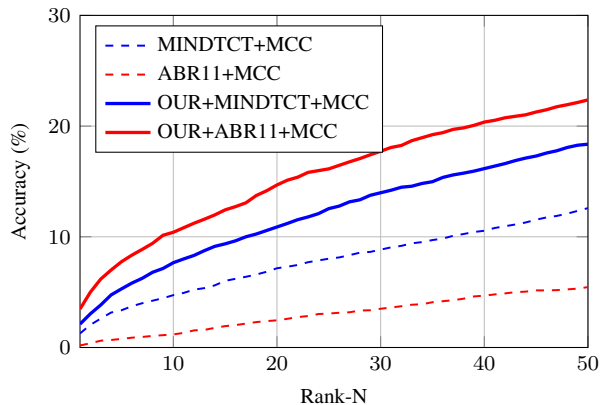

For this experiment, we refer to the case from \cite{Sankaran15}, where minutiae were extracted automatically and afterwards matched using one of the standard algorithms. Sankaran \etal evaluated the performance of publicly available NBIS and commercial VeriFinger \cite{VeriFinger} fingerprint matching methods, reporting very poor performance for both. Here, we used the ABR11 + MCC method for matching. Table \ref{tab:LatentToSensor} shows that using our enhancement algorithm, ABR11 + MCC performs very poorly, consistently with the finding of \cite{Sankaran15}. Performing the enhancement with our model significantly improves the performance. CMC curves for this experiment are shown Figure \ref{fig:cmcLatentToSensor}.


\begin{table}[!t]
\begin{center}
\begin{tabular}{cccc}
\multicolumn{2}{c}{Method} & \multicolumn{2}{c}{Accuracy} \\
  \cline{1-2}\cline{3-4}
Enhancement & Extract + Match & Rank-1 & Rank-10 \\
\hline
Raw & ABR11 + MCC & 35.58\% & 58.27\% \\
Raw & NBIS & 57.69\% & 71.92\%  \\
Our & ABR11 + MCC & 57.12\% & 79.04\%   \\
Our & NBIS & 62.69\% & 78.85\%  \\
\hline
\end{tabular}
\end{center} 
\caption{ Latent-to-Latent matching on IIIT-Delhi latent database. Matching was performed on images obtained with the proposed fingerprint enhancement (Our) and with no enhancement (Raw). }
\label{tab:LatentToLatent}
\end{table}

 \begin{table}[!t]
\begin{center}
\begin{tabular}{cccc}
\multicolumn{2}{c}{Method} & \multicolumn{2}{c}{Accuracy} \\
  \cline{1-2}\cline{3-4}
Enhancement & Extract + Match & Rank-25 & Rank-50 \\
\hline
Raw & MINDTCT + MCC & 8.02\% & 12.59\% \\
Raw & ABR11 + MCC & 3.07\% & 5.45\% \\
Our & MINDTCT + MCC & 12.55\% & 18.36\% \\
Our & ABR11 + MCC & 16.14\% & 22.36\%   \\
\cite{Sankaran15} & NBIS & N/A & 6.06\% \\
\cite{Sankaran15} & VeriFinger & N/A & 6.80\% \\
\hline
\end{tabular}
\end{center}
\caption{ Latent-to-Sensor matching on IIIT-Delhi MOLF database. Shown as the performance of matching Lumidigm images to latent fingerprints with enhancement (Our) and with no enhancement (Raw). Two more methods listed in \cite{Sankaran15} are compared. }
\label{tab:LatentToSensor}
\end{table}
 
\begin{figure}[!h]
\centering
\setlength\figureheight{6cm} 
\setlength\figurewidth{\linewidth}
%
%
%
\begin{tikzpicture}

\pgfplotsset{compat=newest} 

\definecolor{color0}{rgb}{1, 1, 1}

\tikzstyle{every node}=[font=\footnotesize]

\begin{axis}[
xlabel={Rank-N},
ylabel={Accuracy (\%)},
xmin=1, xmax=50,
ymin=0, ymax=30,
width=\figurewidth,
height=\figureheight,
at={(0\figurewidth,0\figureheight)},
xmajorgrids,
x grid style={lightgray},
ymajorgrids,
y grid style={lightgray},
axis line style={black},
axis background/.style={fill=color0},
legend style={at={(0.03,0.97)}, anchor=north west},
legend cell align={left},
legend entries={{Raw-Crossmatch},{Raw-Secugen},{Raw-Lumidigm},{Our-Crossmatch},{Our-Secugen},{Our-Lumidigm}}
]
\addplot [dashed, line width=0.75pt, green]
table {%
    1.0000    0.2273
    2.0000    0.3409
    3.0000    0.5455
    4.0000    0.6591
    5.0000    0.7727
    6.0000    0.8636
    7.0000    0.9773
    8.0000    1.0682
    9.0000    1.1818
   10.0000    1.3182
   11.0000    1.3636
   12.0000    1.4318
   13.0000    1.5227
   14.0000    1.6136
   15.0000    1.7727
   16.0000    1.8864
   17.0000    2.0000
   18.0000    2.0682
   19.0000    2.1591
   20.0000    2.1818
   21.0000    2.2045
   22.0000    2.3864
   23.0000    2.5227
   24.0000    2.6364
   25.0000    2.6818
   26.0000    2.7727
   27.0000    2.8636
   28.0000    3.0000
   29.0000    3.1591
   30.0000    3.3182
   31.0000    3.4545
   32.0000    3.5909
   33.0000    3.6364
   34.0000    3.6818
   35.0000    3.7273
   36.0000    3.7955
   37.0000    3.8409
   38.0000    3.9545
   39.0000    4.0227
   40.0000    4.1591
   41.0000    4.2500
   42.0000    4.3409
   43.0000    4.5455
   44.0000    4.7955
   45.0000    4.8636
   46.0000    4.8864
   47.0000    4.9318
   48.0000    5.0000
   49.0000    5.1591
   50.0000    5.3182
};
\addplot [dashed, line width=0.75pt, blue]
table {%
    1.0000    0.2500
    2.0000    0.4091
    3.0000    0.5909
    4.0000    0.6591
    5.0000    0.7955
    6.0000    0.8409
    7.0000    0.9318
    8.0000    1.0000
    9.0000    1.1136
   10.0000    1.2273
   11.0000    1.3864
   12.0000    1.4773
   13.0000    1.5909
   14.0000    1.6136
   15.0000    1.7273
   16.0000    1.8409
   17.0000    1.9545
   18.0000    2.0909
   19.0000    2.1136
   20.0000    2.1591
   21.0000    2.2727
   22.0000    2.2955
   23.0000    2.4318
   24.0000    2.5227
   25.0000    2.5909
   26.0000    2.6818
   27.0000    2.8864
   28.0000    3.0227
   29.0000    3.1591
   30.0000    3.2273
   31.0000    3.3636
   32.0000    3.5455
   33.0000    3.6591
   34.0000    3.8409
   35.0000    3.9545
   36.0000    4.0455
   37.0000    4.0682
   38.0000    4.1364
   39.0000    4.2273
   40.0000    4.2727
   41.0000    4.3182
   42.0000    4.4091
   43.0000    4.5000
   44.0000    4.6364
   45.0000    4.7273
   46.0000    4.7955
   47.0000    4.9318
   48.0000    5.0000
   49.0000    5.0909
   50.0000    5.1818
};
\addplot [dashed, line width=0.75pt, red]
table {%
    1.0000    0.1818
    2.0000    0.3864
    3.0000    0.6136
    4.0000    0.6818
    5.0000    0.7727
    6.0000    0.8864
    7.0000    0.9545
    8.0000    1.0455
    9.0000    1.1136
   10.0000    1.1818
   11.0000    1.3182
   12.0000    1.5455
   13.0000    1.5909
   14.0000    1.7727
   15.0000    1.9318
   16.0000    2.0227
   17.0000    2.1591
   18.0000    2.2955
   19.0000    2.3864
   20.0000    2.4545
   21.0000    2.5909
   22.0000    2.7500
   23.0000    2.8409
   24.0000    3.0227
   25.0000    3.0682
   26.0000    3.1364
   27.0000    3.1818
   28.0000    3.3409
   29.0000    3.3636
   30.0000    3.5000
   31.0000    3.6136
   32.0000    3.7273
   33.0000    3.8636
   34.0000    3.9091
   35.0000    4.0909
   36.0000    4.1818
   37.0000    4.2727
   38.0000    4.4091
   39.0000    4.6136
   40.0000    4.6818
   41.0000    4.7955
   42.0000    4.8864
   43.0000    5.0000
   44.0000    5.0455
   45.0000    5.1591
   46.0000    5.1591
   47.0000    5.1818
   48.0000    5.2500
   49.0000    5.3182
   50.0000    5.4545
};
\addplot [line width=1.25, green!50.0!black]
table {%
    1.0000    2.5682
    2.0000    3.6364
    3.0000    4.5682
    4.0000    5.2500
    5.0000    5.8409
    6.0000    6.3409
    7.0000    6.7500
    8.0000    7.2500
    9.0000    7.6364
   10.0000    7.9091
   11.0000    8.3636
   12.0000    8.6364
   13.0000    8.8182
   14.0000    9.0682
   15.0000    9.3182
   16.0000    9.7273
   17.0000   10.0909
   18.0000   10.4318
   19.0000   10.6591
   20.0000   10.9545
   21.0000   11.2500
   22.0000   11.5909
   23.0000   11.9545
   24.0000   12.2273
   25.0000   12.6591
   26.0000   12.8636
   27.0000   13.1818
   28.0000   13.5227
   29.0000   13.8182
   30.0000   14.0455
   31.0000   14.3182
   32.0000   14.6591
   33.0000   15.0227
   34.0000   15.2500
   35.0000   15.5682
   36.0000   15.8182
   37.0000   16.1364
   38.0000   16.3182
   39.0000   16.6818
   40.0000   16.9091
   41.0000   17.1591
   42.0000   17.3636
   43.0000   17.5000
   44.0000   17.7500
   45.0000   18.0227
   46.0000   18.2500
   47.0000   18.4545
   48.0000   18.7045
   49.0000   18.8636
   50.0000   19.0682
};
\addplot [line width=1.25, blue]
table {%
    1.0000    2.5682
    2.0000    3.4545
    3.0000    4.3864
    4.0000    5.0682
    5.0000    5.6364
    6.0000    6.2955
    7.0000    6.7273
    8.0000    7.3636
    9.0000    7.9091
   10.0000    8.2500
   11.0000    8.6818
   12.0000    9.0682
   13.0000    9.3636
   14.0000    9.7045
   15.0000   10.1591
   16.0000   10.5682
   17.0000   10.9545
   18.0000   11.2500
   19.0000   11.4545
   20.0000   11.8636
   21.0000   12.0909
   22.0000   12.3864
   23.0000   12.6364
   24.0000   12.9091
   25.0000   13.2727
   26.0000   13.7500
   27.0000   13.9545
   28.0000   14.1591
   29.0000   14.6136
   30.0000   14.8409
   31.0000   15.0682
   32.0000   15.3409
   33.0000   15.6364
   34.0000   15.8182
   35.0000   16.1364
   36.0000   16.4091
   37.0000   16.7273
   38.0000   16.9773
   39.0000   17.2273
   40.0000   17.4318
   41.0000   17.7955
   42.0000   18.0455
   43.0000   18.2955
   44.0000   18.4773
   45.0000   18.6591
   46.0000   18.7727
   47.0000   18.9091
   48.0000   19.1818
   49.0000   19.2727
   50.0000   19.5000
};
\addplot [line width=1.25, red]
table {%
    1.0000    3.4773
    2.0000    5.0000
    3.0000    6.1818
    4.0000    7.0000
    5.0000    7.7500
    6.0000    8.3409
    7.0000    8.8636
    8.0000    9.4091
    9.0000   10.1364
   10.0000   10.4091
   11.0000   10.8182
   12.0000   11.2273
   13.0000   11.5909
   14.0000   11.9773
   15.0000   12.4318
   16.0000   12.7273
   17.0000   13.0682
   18.0000   13.7273
   19.0000   14.1591
   20.0000   14.6818
   21.0000   15.1136
   22.0000   15.3864
   23.0000   15.8182
   24.0000   15.9773
   25.0000   16.1364
   26.0000   16.4545
   27.0000   16.7727
   28.0000   17.0682
   29.0000   17.4091
   30.0000   17.7273
   31.0000   18.0682
   32.0000   18.2500
   33.0000   18.6818
   34.0000   18.9545
   35.0000   19.2273
   36.0000   19.4091
   37.0000   19.7045
   38.0000   19.8409
   39.0000   20.0682
   40.0000   20.3636
   41.0000   20.5227
   42.0000   20.7500
   43.0000   20.8864
   44.0000   21.0227
   45.0000   21.2727
   46.0000   21.4773
   47.0000   21.7500
   48.0000   21.9318
   49.0000   22.1364
   50.0000   22.3636
};
\end{axis}

\end{tikzpicture}
\caption{Cross-dataset Latent-to-Sensor matching on IIIT-Delhi MOLF database. CMC curve showing the performance improvement of our enhancement approach (solid lines) against matching raw latent fingerprints (dashed lines) to the Lumidigm, Secugen, and Crossmatch scanner datasets. }
\label{fig:cmcLatentToSensorDifferent}
\end{figure}
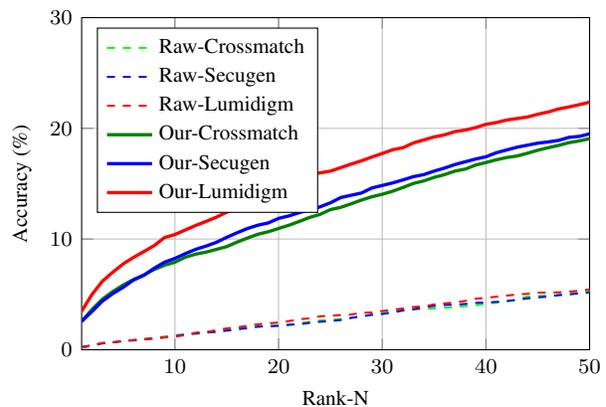

\begin{table}[!h]
\begin{center}
\begin{tabular}{cccc}
\multicolumn{2}{c}{Method} & \multicolumn{2}{c}{Accuracy} \\
  \cline{1-2}\cline{3-4}
Enhancement & Gallery dataset & Rank-25 & Rank-50 \\
\hline
Raw & Lumidigm & 3.07\% & 5.45\% \\
Raw & Secugen & 2.59\% & 5.18\% \\
Raw & Crossmatch & 2.68\% & 5.32\% \\
Our & Lumidigm & 16.14\% & 22.36\%   \\
Our & Secugen & 13.27\% & 19.50\%   \\
Our & Crossmatch & 12.66\% & 19.07\%   \\
\hline
\end{tabular}
\end{center}
\caption{ Cross-dataset Latent-to-Sensor matching on IIIT-Delhi MOLF database. Performance of matching Lumidigm, Secugen, and Crossmatch sensor images to latent fingerprints enhanced by our model (Our) and with no enhancement (Raw). In all cases, combination ABR11 + MCC was used for feature extraction and matching. }
\label{tab:LatentToSensorDifferent}
\end{table}

\paragraph*{\bf Cross-dataset Latent-to-Sensor matching. }
To support the fact that our method works not only matching latent fingerprints to Lumidigm (MOLF DB1) sensor scanned fingerprints, we provide a comparison of matching latent fingerprints to Secugen (MOLF DB2) and Crossmatch (MOLF DB3) sensor samples as well using the combination of ABR11 feature extractor and MCC matching method.   
As samples from various sensors are of different quality, the performance can differ slightly. However, Table \ref{tab:LatentToSensorDifferent} indicates that independently of the source sensor, our enhancement algorithm provides significant performance improvement. The CMC curves for this experiment are shown in Figure \ref{fig:cmcLatentToSensorDifferent}.

%
%

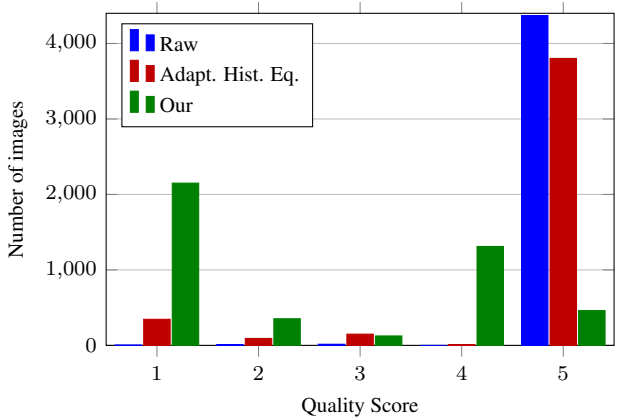
\begin{figure}[!h]
\centering
\setlength\figureheight{6cm} 
\setlength\figurewidth{\linewidth}
%
%
%
\begin{tikzpicture}

\pgfplotsset{compat=newest} 

\definecolor{color0}{rgb}{1, 1, 1}

\tikzstyle{every node}=[font=\footnotesize]

\begin{axis}[
xlabel={Quality Score},
ylabel={Number of images},
xmin=0.5, xmax=5.5,
ymin=0, ymax=4400,
width=\figurewidth,
height=\figureheight,
at={(0\figurewidth,0\figureheight)},
x grid style={lightgray},
ymajorgrids,
ybar=2*\pgflinewidth,
bar width=10pt,
y grid style={lightgray},
axis line style={black},
axis background/.style={fill=color0},
legend style={at={(0.03,0.97)}, anchor=north west},
legend cell align={left},
legend entries={{Raw}, {Adapt. Hist. Eq.}, {Our}}
]
\addplot [style={blue, fill=blue, mark = none}]
  coordinates {(1.0, 4) (2.0, 11) (3.0, 15) (4.0, 0) (5.0, 4370)};
\addplot [style={red!75.0!black, fill=red!75.0!black, mark = none}]
  coordinates {(1.0, 346) (2.0, 92) (3.0, 149) (4.0, 11) (5.0, 3802)};
\addplot [style={green!50.0!black, fill=green!50.0!black, mark = none}]
  coordinates {(1.0, 2150) (2.0, 353) (3.0, 125) (4.0, 1311) (5.0, 461)};
\end{axis}

\end{tikzpicture}
\caption{Image quality assessment using NFIQ utility. We compare the raw data with data improved using adaptive histogram equalization and data improved using our enhancement. The quality assessment score ranges from 1-5, where the lower the better. }
\label{fig:histImageQuality}
\end{figure}

\paragraph*{\bf Fingerprint quality.  }
In order to qualitatively demonstrate how well our model reconstructs the latent fingerprints, we perform quality assessment using the NFIQ utility (part of NBIS \cite{NBIS}), which assigns a fingerprint image a numerical score from 1 (best quality) to 5 (poorest quality). The distribution of score values shown in Figure \ref{fig:histImageQuality} clearly indicates that our method improves the fingerprint quality. 

\begin{figure}
\centering
\subfigure[Successful reconstructions]
{
\includegraphics[width=2.06in]{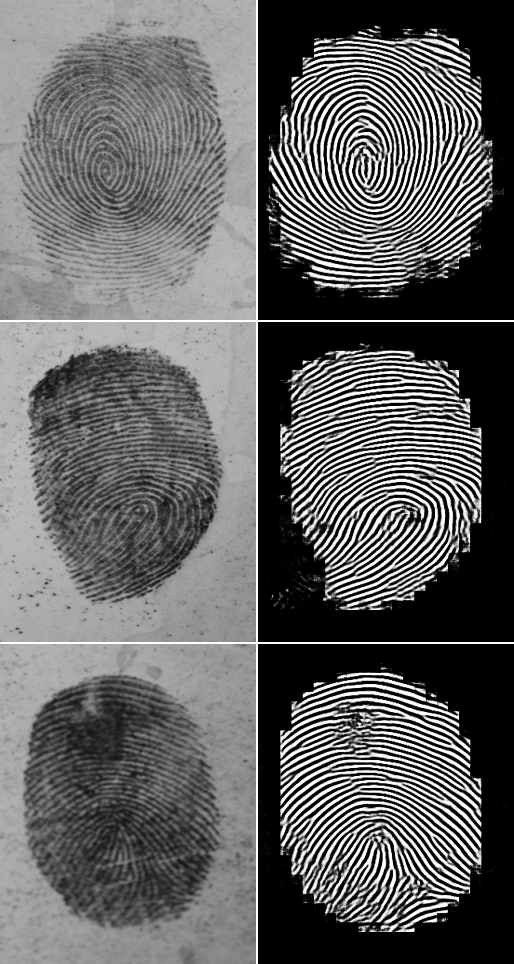}
}
\subfigure[Fail cases]
{
\includegraphics[width=2.06in]{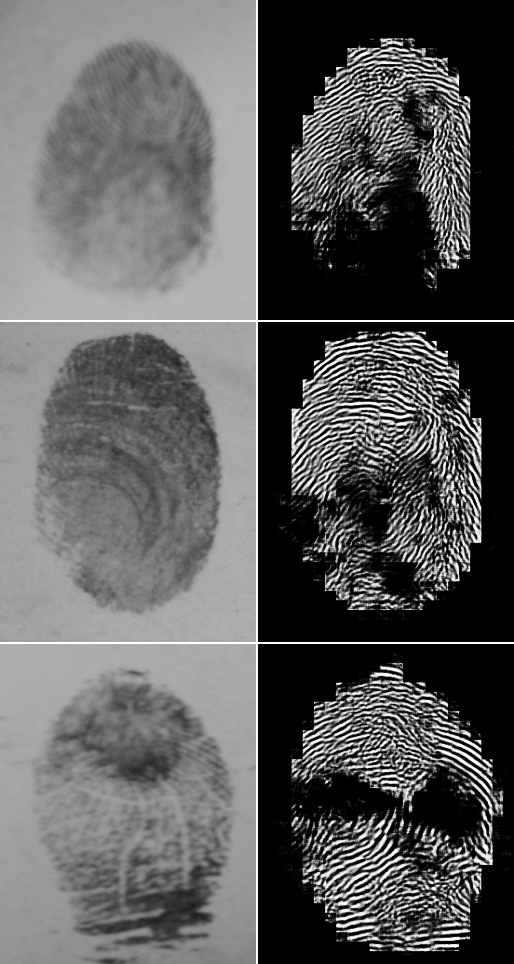}
}
\caption{ Examples of the fingerprint reconstructions on real latent fingerprints. Each pair is (input, reconstruction) (a) successfully reconstructed samples (b) cases where the model fails to reconstruct the fingerprint well. }
\label{fig:ReconstructionExamplesReal}
\end{figure}

\begin{figure*}
\centering
\subfigure[Successful reconstructions]
{
\includegraphics[width=3.3in]{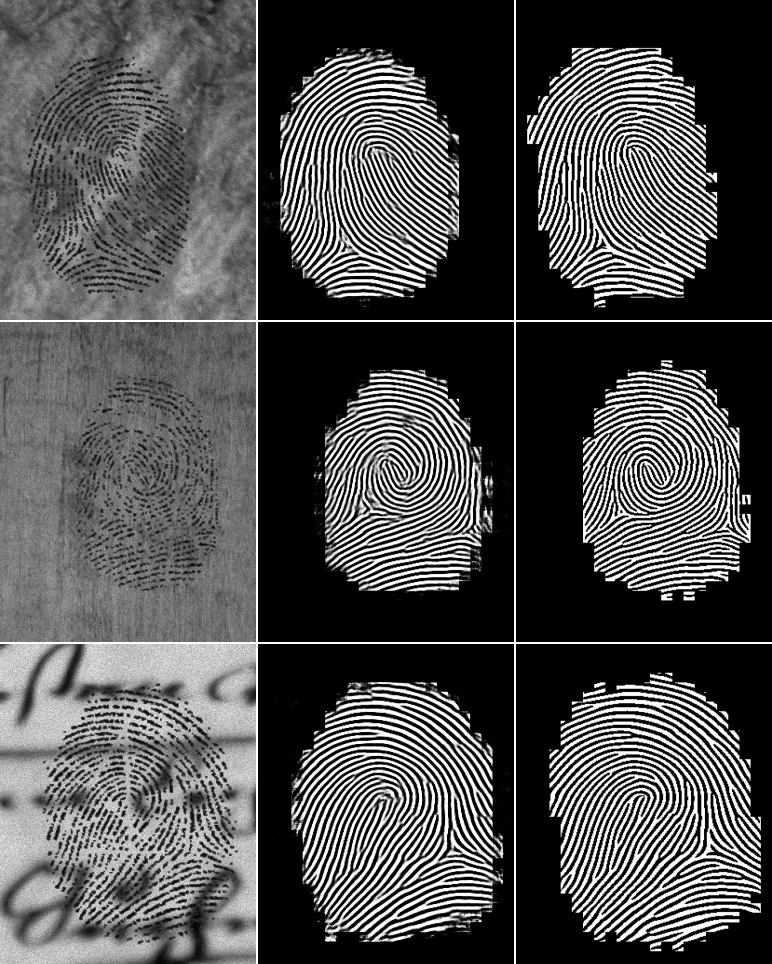}
}
\subfigure[Fail cases]
{
\includegraphics[width=3.3in]{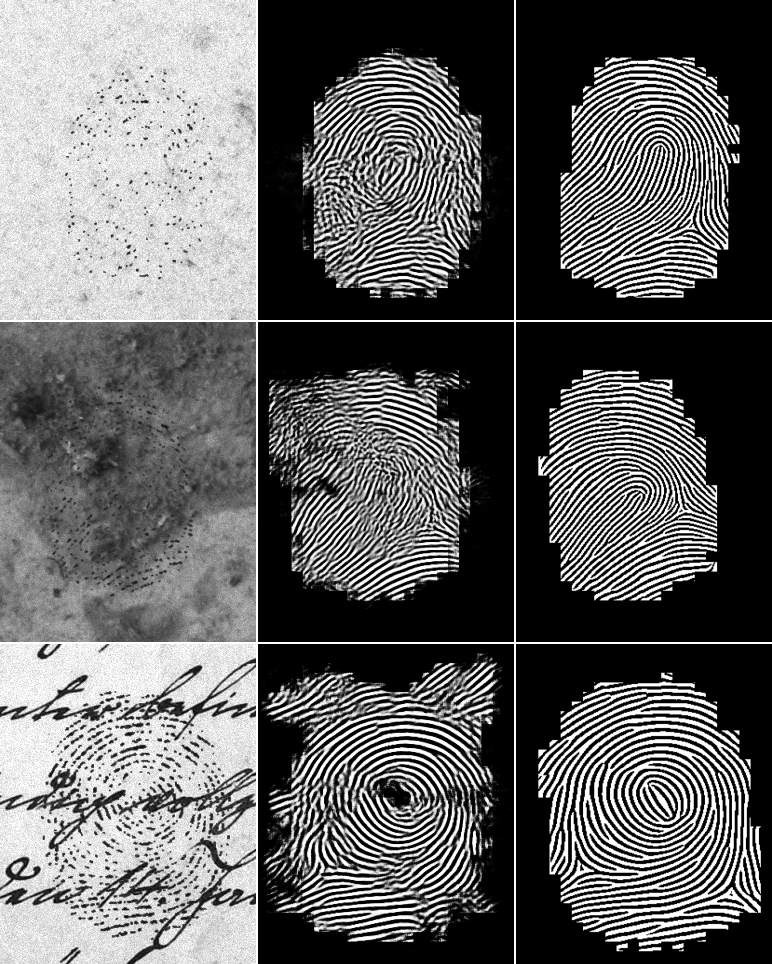}
}
\caption{ Examples of the fingerprint reconstructions on synthetic data. Each triplet is from left to right (input, reconstruction, target) (a) successfully reconstructed samples (b) cases where the model fails to reconstruct the fingerprint well. }
\label{fig:ReconstructionExamples}
\end{figure*}

Moreover, we show several successful and several unsuccessful latent fingerprint reconstructions for both real (Figure \ref{fig:ReconstructionExamplesReal}) and synthetic (Figure \ref{fig:ReconstructionExamples}) data. It is clearly visible that our model enhances the ridge information well when it is present and has problems in places where the original images does not contain discernible ridges. This suggests that very poor samples might not contain any meaningful information for the reconstruction process. 

\section{Conclusion}
Convolutional autoencoders are amongst popular methods extensively used in image processing tasks for image denoising and inpainting. Inspired by the previous applications, we have explored the possibility of using the convolutional autoencoders to reconstruct latent or damaged fingerprints. With regard to principles of some of the fingerprint feature extraction and matching algorithms, we have carefully designed an objective function that should both well reflect on the important fingerprint properties and be efficient to optimize as it is based on gradient analysis, which has been already implemented on the GPUs.
Our method is based on learning, however, in comparison to some of the previous research, we do not need any real training data and we effectively train our model on well designed synthetic dataset, which gives us the advantage of no training set size limitation. 

We obtain state-of-the-art results on several challenging tasks such as latent-to-latent fingerprint matching and latent-to-sensor database fingerprint matching on IIIT-D standard datasets, outperforming the existing results using on the same data by a margin. Especially evaluation on very challenging IIIT-D MOLF dataset compensates for the fact that we cannot evaluate on NIST-SD27 as it is discontinued and no longer provided by NIST.

On the other hand, we observe that the reconstruction is not always successful, showing some of the failure cases that are prone to generate false minutiae. We would like to further examine this issue as future work.

As we do not aim to directly extract minutiae or perform matching, but rather reconstruct the correct ridge pattern of the poor-quality fingerprint, our method has broad applications ranging from latent fingerprint enhancement to possibly reconstruction of fingerprints affected by diseases, which is another of our desired future directions.

\section{Acknowledgements}
The authors are supported in part by ERC Starting Grant No. 307047 (COMET), ERC Consolidator Grant No. 724228 (LEMAN) and Nvidia equipment grant.

{\small
\bibliographystyle{ieee}
\bibliography{bibliography}
}

\end{document}